\documentclass[conference]{IEEEtran}
\IEEEoverridecommandlockouts
\usepackage[linesnumbered,ruled,noline]{algorithm2e}

\usepackage{CJKutf8}

\usepackage[whole]{bxcjkjatype}

\usepackage{url}
\usepackage{color}
\usepackage{cite}
\usepackage{amsmath,amssymb,amsfonts}
\usepackage{algorithmic}
\usepackage{graphicx}
\usepackage{textcomp}
\usepackage{xcolor}
\def\BibTeX{{\rm B\kern-.05em{\sc i\kern-.025em b}\kern-.08em
    T\kern-.1667em\lower.7ex\hbox{E}\kern-.125emX}}
    
\begin{document}

\title{Medical Incident Causal Factors and Preventive Measures Generation Using Tag-based Example Selection in Few-shot Learning\\
}

\makeatletter
\newcommand{\linebreakand}{%
  \end{@IEEEauthorhalign}
  \hfill\mbox{}\par
  \mbox{}\hfill\begin{@IEEEauthorhalign}
}
\makeatother

\author{
\IEEEauthorblockN{Yuna Haseyama}
\IEEEauthorblockA{
\textit{Graduate School of}\\
\textit{Information Science and Technology}\\
\textit{Hokkaido University}\\
Hokkaido, Japan\\
haseyama.yuna.n3@elms.hokudai.ac.jp}
\and

\IEEEauthorblockN{Tomoki Ito}
\IEEEauthorblockA{
\textit{National Institute of Information}\\ 
\textit{and Communications Technology (NICT)}\\
Kyoto, Japan\\
m2015titoh@socsim.org}

\linebreakand

\IEEEauthorblockN{Hiroki Sakaji}
\IEEEauthorblockA{
\textit{Faculty of Information Science and Technology}\\
\textit{Hokkaido University}\\
Hokkaido, Japan\\
sakaji@ist.hokudai.ac.jp}
\and
\IEEEauthorblockN{Itsuki Noda}
\IEEEauthorblockA{
\textit{Faculty of Information Science and Technology}\\
\textit{Hokkaido University}\\
Hokkaido, Japan\\
i.noda@ist.hokudai.ac.jp}
}


\maketitle

\begin{abstract}
In high-stakes domains such as healthcare, the reliability of Large Language Models (LLMs) is critical, particularly when generating clinical insights from incident reports.
This study proposes a tag-based few-shot example selection method for prompting LLMs to generate background/causal factors and preventive measures from details of the medical incidents.
For our experiments, we use the Japanese Medical Incident Dataset (JMID), a structured dataset of 3,884 real-world medical accident and near-miss reports. 
These reports are variably annotated with a wide range of tags—some include descriptive information (e.g., “medications,” “blood transfusion therapy”).
We compare three few-shot example selection strategies—random sampling, cosine similarity-based selection, and our proposed tag-based method—using GPT-4o and LLaMA 3.3. 
Results show that the tag-based approach achieves the highest precision and most stable generation behavior, while similarity-based selection often leads to unintended outputs and safety filter activation. 
These findings suggest that selecting examples based on human-interpretable dataset tags can improve generation precision and stability in clinical LLM applications.

\end{abstract}


\section{Introduction}

\subsection{Motivation and Purpose}
Large Language Models (LLMs), which possess the capability to comprehensively understand and integrate a broad spectrum of medical knowledge, are increasingly being recognized for their potential to enhance operational efficiency and support clinical decision-making in healthcare settings, due to their advanced natural language processing abilities.
According to the findings in \cite{summarization}, clinical summaries generated by LLMs were frequently preferred over those written by clinicians themselves.
This implies that LLMs may offer a viable solution to support clinical documentation tasks, potentially improving workflow efficiency and reducing the administrative burden on healthcare providers.
Moreover, a lack of clarity in patient reports has been linked to a decline in the quality of patient care \cite{patientreports}, highlighting the clear importance of effective communication between patients and healthcare providers. 
Given their ability to simplify complex text and their extensive medical knowledge, LLMs have been suggested as a potential means to improve the quality of such communication \cite{communicate}.
However, LLMs often produce inaccuracies or "hallucinations," which pose a significant barrier to their practical use in clinical settings—environments where errors by healthcare professionals can mean the difference between life and death for patients \cite{hallucination}.
In this study, we focus on medical safety within the healthcare domain and propose a tag-based example selection method for few-shot learning in order to improve precision in the task of generating the background/causal factors and preventive measures from details of the medical incidents descriptions.
In clinical practice, when a medical incident occurs, it is critically important to promptly and accurately identify its causes and implement preventive measures to avoid recurrence.
 To support this, Japan’s Ministry of Health, Labor and Welfare has been implementing the "Project to Collect Medical Adverse Event Information"\footnote{\url{https://www.med-safe.jp/}} since 2004.
 Through this initiative, harmful events such as medical accidents are continuously collected, and reported cases—including near-miss incidents—are analyzed and made publicly available to medical institutions and the general public to provide useful information for improving patient safety.
Despite the fact that patient safety and medical incidents are critical issues within the medical domain, unfortunately, there has been limited research conducted on this subject. 
Building on this context, this paper aims to develop the method of automatically generating the background/causal factors and preventive measures  of each medical incident.
Here, we note the following significant points for this study:
\begin{itemize}
\item First, precision is important for this task because the wrong output can bring other trouble to the point of care.
\item Second, in this paper, we focus on the few-shot learning setting using LLMs because the size of the training dataset is often limited in this area; therefore, a method that can be utilized in such a situation is preferable.
\end{itemize}

\subsection{Our approach}
To achieve this aim, we first prepared a  dataset called Japanese Medical Incident Dataset (JMID), based on reports provided by the Japan Council for Quality Health Care (JQ)\footnote{\url{https://jcqhc.or.jp/},\url{https://jcqhc.or.jp/en/}} for the evaluation.
This dataset includes ''Details of the Incident", ''Background/Causal Factors", and ''Preventive Measures" of each medical incident as shown in Table \ref{tab:dataset}.
In addition, "descriptive information" tag as shown in Table \ref{topic} is assigned to certain incidents. 

We then propose a novel example selection method called  ''tag-based example selection" in few-shot learning setting using LLMs.
Our approach selects few-shot learning examples within the same category 
based on the ''descriptive information" tag that can be available on the dataset (Table \ref{topic}).

In contrast to the random selection approach, our method is expected to prevent the generation of  unrelated content.
In addition, in contrast to the selection method based on cosine similarity between the input and incident descriptions, the examples selected by our method are expected to be more diverse; therefore, the outputs generated using our approach are expected to be more flexible.
\subsection{Contribution}
The main contributions of this study are as follows: 

\begin{itemize} 
    \item Proposal of a tag-based few-shot example selection method: We introduce a method that selects relevant examples based on descriptive tags (e.g., medication, blood transfusion, patient misidentification) from medical incident reports, achieving higher precision than random or similarity-based approaches.
    \item Challenge of patient safety and medical incidents: there has been limited prior work explicitly focusing on automatically generating "Background/Causal Factors" and "Preventive Measures" from medical incident reports.

    This should be a significant challenge in the medical domain.
\end{itemize}

\section{Related Work}
The output performance of LLMs is widely known to vary significantly depending on prompt design, and the technique of prompt engineering has recently emerged as an active area of research \cite{chen2024unleashingpotentialpromptengineering}.
%
Among the various prompting techniques, few-shot learning has gained attention as a method that enables high-quality output using only a small number of examples, without the need for additional training or fine-tuning, and without requiring thousands or tens of thousands of training examples.
In few-shot learning, the model is presented with several input-output pairs in the prompt to help it understand the task context.
In this setting, the selection of which examples to include is of critical importance, as it has been shown to significantly affect model performance \cite{few-shot}.
While early approaches often relied on random selection of examples, it has been reported that such a strategy can still yield competitive results in certain cases \cite{random}. More recent work has explored example selection based on similarity metrics, such as cosine similarity, to identify semantically relevant examples \cite{gpt3}. In contrast to surface-level similarity, Zhang et al. \cite{skillbasedfewshotselectionincontext} proposed a skill-based selection method, which prioritizes examples based on the underlying reasoning skills required by the task.
This approach has demonstrated superior performance compared to traditional similarity-based methods. 
Furthermore, recognizing the limitations of individual strategies, Pecher et al. introduced ACSESS \cite{automaticcombinationsampleselection}, which automatically combines multiple selection strategies and dynamically selects the most effective one per task, yielding consistently higher performance.
In the medical field, numerous tasks have been challenged \cite{medqa}\cite{pubmedqa}\cite{multimedqa}.
In these researches, LLMs are utilized in order to solve the medical Question \& Answering tasks and knowledge and reasoning tasks \cite{medgemini}\cite{meditron}\cite{medpalm2}.
In Japanese, similar efforts have been made, including IgakuQA\cite{igakuqa}, a QA task based on past questions from Japan's national medical licensing examinations from 2018 to 2022.
In addition, several medical tasks including information extraction task and QA task related to medicine and disease have been challenged using Large Language Models in JMED-LLM project\footnote{\url{https://github.com/sociocom/jmed-llm}}.
%

However, to the best of our knowledge, there is still limited research specifically focused on the generation task of medical incidents and patient safety.
%
In this study, we focus on the medical domain—specifically the field of patient safety, where mistakes can have critical consequences—to examine how differences in few-shot example selection impact the precision of model outputs.
We propose a novel selection method that uses the JMID dataset and leverages the "description tags" annotated within each case as the basis for selecting examples.
We argue that such an approach is particularly meaningful in the medical domain, where the availability of training data is limited due to privacy constraints.
Moreover, since our method is based on pre-existing categorical tags, it offers practical advantages: it incurs minimal computational cost and can be easily integrated with existing databases.

\section{Tag-based Example Selection}
This section describes our proposed approach called ''tag-based example selection''.

First, we classify the data according to the “descriptive information” tags contained in the dataset.
Categories with a small amount of data were manually consolidated by merging finer-grained labels into broader categories, leading to a total of 18 categories in this study.
For instance, “Medications” and “Other Medications” were merged into “Medications”, and “Clinical Tests” and “Laboratory Tests” were merged into “(Clinical) Laboratory Tests”. 
The complete mapping of finer-grained labels to the broader categories is provided in Appendix A (Table~\ref{map}), and the list of the 18 category tags is shown in Table \ref{tab:topic}.

Based on these tags, we then randomly select five examples from within each category. 
The selected examples are subsequently used as few-shot instances. 
The pseudocode of the procedure is presented in Algorithm \ref{algorithm}, and the prompt used in our approach is shown in Appendix B (Table \ref{tab:prompt}).

In contrast to the random selection approach, our method is expected to prevent to generate the unrelated contents.
In addition, in contrast to the selection method using cosine similarity between the incident descriptions, the examples selected by our method is 
expected to be diverse; therefore, the generated outputs using our approach is expected to be more flexible.

\begin{table}[]
    \centering
    \small
    \caption{The contents and number of instances for 18 categories, which were manually classified based on the 'descriptive information' tag in the dataset, are described below.}
    \label{tab:topic}
    \renewcommand{\arraystretch}{1.4}
    \begin{tabular}{|l|}
    \hline
1. Dispensing (71 cases)\\
2. Medications (438 cases)\\
3. Mechanical Ventilators (182 cases)\\
4. Infusion Pumps (114 cases)\\
5. Patient-Brought Medications (136 cases)\\
6. Drain Insertion and Management (134 cases)\\
7. (Clinical) Laboratory Tests (84 cases)\\
8. Pediatric Patient Care (133 cases)\\
9. Rehabilitation (131 cases)\\
10. Blood Transfusion Therapy (121 cases)\\
11. Contraindicated Drugs (83 cases)\\
12. Radiological Examinations (71 cases)\\
13. Left-Right (Body Part) Confusion (73 cases)\\
14. Retained Foreign Objects (23 cases)\\
15. Hospital Room Equipment (e.g., beds) (71 cases)\\
16. Chemotherapy (60 cases)\\
17. Electrosurgical Units and Similar Devices (39 cases)\\
18. Patient Misidentification (53 cases)\\
    \hline
    \end{tabular}
    \label{topic}
\end{table}

\begin{algorithm}[ht!]
\label{algorithm}
\SetAlgoNoLine
\KwIn{Dataset $\mathcal{D}$, Target topic $T$}
\KwOut{Sample dataset $\mathcal{S}$}
\If{$T \notin \mathrm{CACHE}$}{
    $\mathcal{L} \leftarrow \emptyset$\;
    \ForEach{$d \in \mathcal{D}$}{
        \If{$d[\text{tag}] = T$}{
            $\mathcal{L} \leftarrow \mathcal{L} \cup \{d[\text{text}]\}$\;
        }
    }
    $\mathrm{CACHE}(T) \leftarrow \mathcal{L}$\;
}
$\mathcal{S} \leftarrow \mathrm{get\_random}(\mathrm{CACHE}(T), 5)$\;
\caption{
Sample selection algorithm that retrieves five random examples for a target topic $T$ from dataset $\mathcal{D}$. \\
$\mathcal{D}$ is the input dataset;\\
$T$ is the target topic (i.e., Dispensing);\\ 
$\mathcal{L}$ is a temporary list to store matched texts; \\
$\mathrm{CACHE}$ is a cache to store previous results; \\
$\mathcal{S}$ is the output sample set;\\ 
$\mathrm{get\_random}(\cdot, 5)$ randomly selects five items from the input list.
}
\end{algorithm}

\section{Experimental Evaluation}
This section eveluates the proposed method using real textual dataset.
\subsection{Japanese Medical Incident Dataset (JMID)}
\begin{table*}
            \centering
            \small
            \caption{Example from the Created Dataset (7th Report)}
            \renewcommand{\arraystretch}{1.4}
            \begin{tabular}{|p{4.3cm}|p{4.3cm}|p{4.3cm}|p{1.5cm}|p{1cm}}
            \hline
                \textbf{Details of the Incident} & \textbf{Background/Causal Factors} &  \textbf{Preventive Measures} &\textbf{Descriptive\newline information} & \dots \\ 
            \hline
                 注射箋「ソル・メドロール２５０mg＋生食１００ml」の処方であったがソル・メドロール１２５mgの溶解液１本にエクサシン１Ａを入れてしまった。\newline
                 (Although the prescribed injection was “Sol-Medrol 250 mg + 100 ml normal saline,” one vial of Sol-Medrol 125 mg diluent was mistakenly mixed with one ampule of Exacin.)
            &
                業者から薬剤をもらい、至急調剤し病棟にあげなくてはならなかったという状況と、慌てた気持ちがあった。ソル・メドロールの置場の小箱に間違ってエクサシンが入っていた。\newline
                (There was a situation in which the medication had to be received from the supplier, prepared urgently, and delivered to the ward, which created a sense of haste. In addition, Exacin had been mistakenly placed in the small box where Sol-Medrol was stored.)
            &
                ・調剤時は慌てずに行う。\newline
                ・返品の薬を戻すとき充分注意が必要であるため、職場で再度統一し、必要であればダブルチェックなど行う。\newline
                (・Dispensing should be carried out without haste.\newline
                ・When returning medications, it is necessary to exercise sufficient caution. Therefore, procedures should be standardized again in the workplace, and, if necessary, double checks should be performed.)
            &
            調剤\newline
            (Dispensing)
            & \dots \\
            
            \hline
                 $\vdots$ & $\vdots$ & $\vdots$ &
                 $\vdots$ &
                 $\ddots$
            \end{tabular}
    \label{tab:dataset}
\end{table*}

%
To evaluate our approch, we use the Japanese Medical Incident Dataset (JMID), which was created based on reports collected and published through the Medical Accident Information Collection Project.
%
This dataset was constructed by extracting tabular information from publicly available PDF reports provided by the Japan Council for Quality Health Care (JQ) as part of the Project to Collect Medical Near-Miss/Adverse Event Information\footnote{\url{https://www.med-safe.jp/contents/report/index.html}}\footnote{\url{https://www.med-safe.jp/contents/english/index.html}}.
These reports compile cases of medical accidents and near misses, and the extracted data includes entries under the headings “Summary of Incident,” “Background and causal factors,” and “preventive measures.” 

The reports were originally written and submitted by healthcare professionals (physicians and nurses).
To address privacy concerns, all personal identifiers have been removed, and the dataset contains no information that could be used to identify individuals.

The extracted data was then converted into jsonl format. 
%
This dataset includes 2,017 entries that contain \textit{descriptive information tags},
which are sourced from publicly available reports published by the Japan Council
for Quality Health Care as part of its Medical Adverse Event Information Collection Project.

The entire JMID is in Japanese, but has been translated in this paper by the authors for readability.

The scope of cases to be reported as medical accident information is specified in the PDF and is as follows.




\begin{enumerate}
    \item[1] Cases in which it is evident that erroneous medical treatment or management was performed, and as a result of such treatment or management, the patient died, suffered physical or mental impairment, or required unexpected or more extensive procedures or other treatments than anticipated.
    \item[2] Cases in which it is not evident that erroneous medical treatment or management was performed, but as a result of such treatment or management, the patient died, suffered physical or mental impairment, or required unexpected or more extensive procedures or other treatments than anticipated (including cases suspected to be caused by the treatment or management performed, limited to those where the occurrence of the incident was not anticipated).
    \item[3] In addition to the cases listed in (1) and (2), cases that contribute to the prevention of accidents within medical institutions and the prevention of recurrence.
\end{enumerate}

Additionally, the information requested for reporting as near-miss incidents includes the following types of cases.
\begin{enumerate}
    \item [1] Cases in which an erroneous medical procedure or similar action was detected before being performed on the patient.
    \item [2] Cases in which an erroneous medical procedure or similar action was performed but did not ultimately affect the patient.
    \item [3] Cases in which an erroneous medical procedure or similar action was performed, resulting in the need for minor procedures or treatment.
\end{enumerate}

An example of this dataset is shown in Table \ref{tab:dataset}.
%
In addition to the fields mentioned above, the dataset includes several tags such as “Flow of Information,” which indicates the timing of the incident, and “Category,” which describes the relevant medical domain or area to which the case pertains.
%

The dataset is publicly available at \url{https://huggingface.co/datasets/sakaji/JMID}.


\subsection{Evaluation}
To evaluate our approach, we assess how well LLMs, in a few-shot learning setting with five examples, generate (1) background/causal factors and (2) preventive measures from medical incident details.
\subsubsection{Comparative Methods}
In order to evaluate the performance of our approach, we compared performance of the following three types of baselines with our approach.
\begin{description}
    \item[\textit{zero-shot (baseline):}]\hspace{6em}
    This method generates the Background/Causal Factors and Preventive Measures using each LLM through the zero-shot learning setting.
    \item[\textit{random(baseline):}]\hspace{6em}
    This method generates the Background/Causal Factors and Preventive Measures using few-shot  (n=5) learning setting where few-shot examples are randomly selected from the entire dataset, and generation is performed using these fixed examples.

\item[\textit{similarity(baseline):}] \hspace{6em}
This method generates the Background/Causal Factors and Preventive Measures using few-shot  (n=5) learning setting where few-shot examples are selected by identifying the five most similar cases to the input based on cosine similarity from the entire dataset, and generation is performed using these examples.
For this setting, the top 5 most similar examples were selected for each input based on the highest cosine similarity scores.
The embedding model used was sentence-luke-japanese-base-lite\footnote{\url{https://huggingface.co/sonoisa/sentence-luke-japanese-base-lite}}, which was chosen due to its reported quantitative performance—comparable to, or slightly better than, Japanese Sentence-BERT models, with a difference of approximately 0 to 0.5 points.
\end{description}

\begin{description}
    \item[\textit{\textbf{tag-based (Our Approach):}}]\hspace{9em}
    This is the method of using the proposed tag-based example selection in a few-shot learning. 
    Here, five examples were randomly sampled from within the tag-based category as the input instance.
    %
    %
    Similar to the \textit{random} setting, once the examples were selected, they were fixed and used consistently across all relevant inputs during the experiment.
\end{description}    

\begin{figure}
    \centering
    \includegraphics[width=1\linewidth]{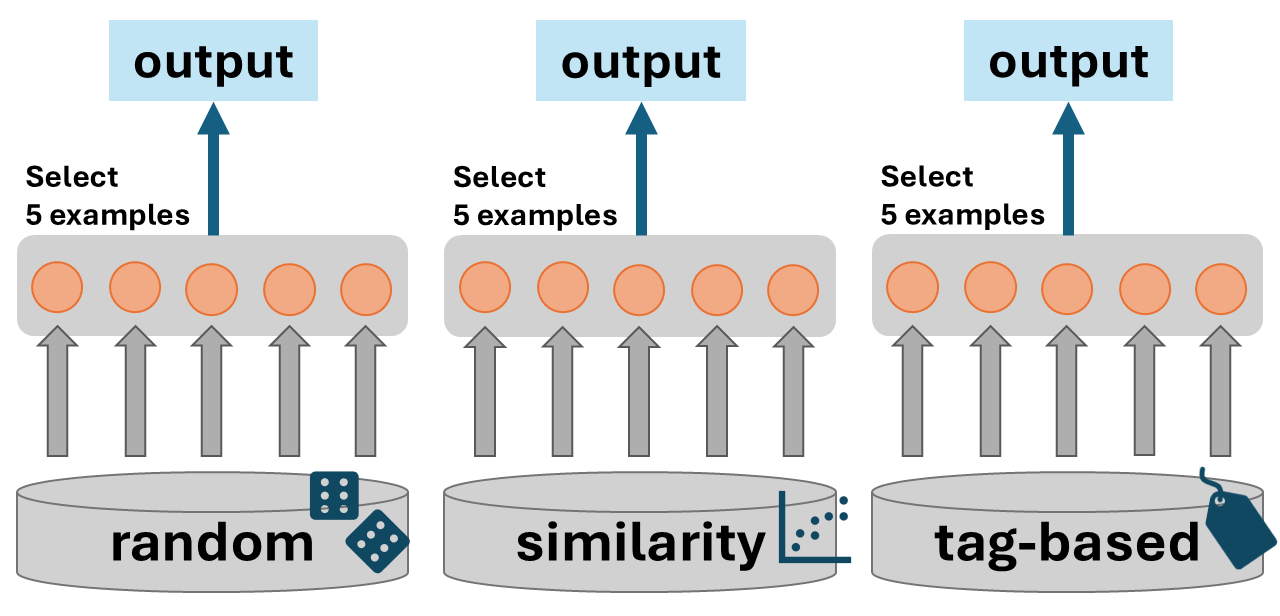}
    \caption{Three example selection methods for few-shot learning: random, similarity, and tag-based. ''tag-based" is the proposed approach.}
    \label{fig:select_examples}
\end{figure}


In this evaluation, we used GPT-4o\footnote{\url{https://openai.com/index/hello-gpt-4o/}} (API version 2024-02-01, context window 128k tokens, accessed via the OpenAI API) and LLaMA 3.3\footnote{\url{https://www.llama.com/}} (70.6 billion parameters, context 
window 131{,}072 tokens, executed with Ollama on an NVIDIA RTX A6000 with 48 GB memory) as LLMs. 
For both models, we fixed the generation parameter temperature to 0 in order to ensure deterministic outputs and reproducibility across experiments, and also to prevent performance variance from sampling randomness that could obscure the effects of few-shot example selection.

As an evaluation metric, we employed ROUGE-1, ROUGE-L \cite{rouge} and BERTScore\footnote{\url{https://github.com/Tiiiger/bert_score}} \cite{bertscore}, using the bert-base-multilingual-cased model\footnote{\url{https://huggingface.co/google-bert/bert-base-multilingual-cased}} \cite{bert_model}. 
ROUGE is an n-gram-based evaluation metric, whereas BERTScore measures semantic similarity between texts. 
In particular, according to Qiu et al. \cite{towards}, in the context of MMedBench\footnote{\url{https://henrychur.github.io/MultilingualMedQA/}}, which is a multilingual benchmark designed for evaluating medical text generation, BERTScore has demonstrated strong alignment with human judgment, indicating its usefulness even in the medical domain. 
By employing both metrics, we are able to capture lexical overlap and semantic similarity, thereby enabling a more balanced and comprehensive evaluation.

\section{Results and Discussion}
\subsection{Results}
Tables \ref{tab:all_cause_ave} and \ref{tab:all_improve} show the evaluation results of each method. 
These results indicate that our approach achieves higher ROUGE-L precision and BERTScore precision than the other baselines.

Here, any outputs that were blocked by GPT-4o's violence filters or failed to follow the expected pattern (such as responding to all five few-shot examples instead of only the sixth prompt, repeating words unnecessarily, or otherwise deviating from the intended format) were assigned a score of zero in the evaluation.
In the \textbf{zero-shot} setting, GPT-4o failed to generate outputs for 12 cases due to its violence filters. In the \textbf{similarity} setting, 19 cases were blocked by the violence filter. In the \textbf{tag-based} setting, all outputs were successfully generated, but 4 of them exhibited unintended behavior.
For Llama 3.3, 3 cases resulted in unintended outputs in both the \textbf{zero-shot} and \textbf{tag-based} settings, while 207 unintended outputs were observed in the \textbf{similarity} setting.
As shown in Tables \ref{tab:all_cause_ave} and \ref{tab:all_improve}, for both the generated background/causal factors and preventive measures, GPT-4o and Llama 3.3 achieve the highest BERTScore and ROUGE-L precision when using the \textbf{tag-based}, followed by the \textbf{random} setting.
Table~\ref{tab:average_similarity} shows the average cosine similarities for each setting.

\begin{table*}[]
    \centering
    \caption{Average scores of background/causal factors generated with GPT-4o and LLaMA 3.3 across zero-shot, random, similarity and tag-based settings.}
    {
    \renewcommand{\arraystretch}{1.2}

    \begin{tabular}{l|ccc|ccc|ccc}
    \hline
    &&BERTScore&&&ROUGE-1&&&ROUGE-L\\
    LLM (method) &Prec. & Rec. & F1&Prec. & Rec. & F1& Prec. & Rec. & F1\\
    \hline
    GPT-4o(zero-shot)
    & .620& .698& .656 
    &.184 &.573&.256
    &.122& .417& .173
\\
    Llama3.3(zero-shot) &.613& .701& .653 
    &.170 &.593&.241
    &.109&.423&.158\\
    \hline
    GPT-4o(random)
    &.707&.717&.711
    &.385& .447& .370
    &.266&.328&.260\\ 
    Llama3.3(random)
    &.690&.715&.701
    &.344&.464& .351&
    .224& .325& .233\\
    \hline
    GPT-4o(similarity)
    &.706&.710&.707
    &.391&.398&.353
    &.236&.253&.216
    
\\
    Llama3.3(similarity)
    &.629&.638&.633
    &.335&.361& .303
    &.231&.260&.210\\

    \hline
    \hline
    GPT-4o(tag-based)
    &\textbf{.712}&.713&.712 
    &\textbf{.402}& .416& .360
    &\textbf{.283}&.309&.257
\\
    Llama3.3(tag-based)
    &\textbf{.703}& .710& .706&
    \textbf{.387}&.398&.342
    &\textbf{.269}&.290& .240\\

    \hline
    \end{tabular}
    }
    \label{tab:all_cause_ave}
\end{table*}


\begin{table*}[]
    \centering
    \caption{Average scores of preventive measures generated with GPT-4o and LLaMA 3.3 across zero-shot, random, similarity and tag-based settings.}
    {
    \renewcommand{\arraystretch}{1.2}

    \begin{tabular}{l|ccc|ccc|ccc}
    \hline
    & & BERTScore && & ROUGE-1 & && ROUGE-L &\\
    LLM (method)&Prec. & Rec. & F1&Prec. & Rec. & F1& Prec. & Rec. & F1\\
    \hline
    GPT-4o(zero-shot) 
    &.609& .698& .650 
    & .106 &.572 &.169
    &.074& .431& .121
\\
    Llama3.3(zero-shot) 
    &.616& .702& .656 
    & .115& .543&.178
    &.079&.403&.124\\
    \hline
    GPT-4o(random)
    &.692&.730&.710 
    &.265& .540 &.329
    &.189& .412& .239
    \\
    Llama3.3(random)
    &.695&.732&.712
    &.280& .532& .338
    &.193&.391&.237\\
    \hline
    GPT-4o(similarity)
    &.696&.725&.710
    &.299&.525&.351
    &.175&.329&.209
\\
    Llama3.3(similarity)
    &.632&.658&.644
    &.279& .452&.314
    &.195&.334&.223\\
    \hline
    \hline
    GPT-4o(tag-based)
    &\textbf{.699}&.729&.713
    &.291& .501& .339
    &\textbf{.297}&.379&.246
    \\
    Llama3.3(tag-based)
    &\textbf{.704}&.730&.716
    &\textbf{.321}&.491&.353
    &\textbf{.223}&.361&.249\\
    \hline
    \end{tabular}
    }
    \label{tab:all_improve}
\end{table*}

\begin{table}[]
    \centering
    \caption{Average Cosine Similarity Between Input and Their Five Few-Shot Examples by Selection Method}
    \resizebox{0.35\textwidth}{!}{%
    \renewcommand{\arraystretch}{1.3}

    \begin{tabular}{cc}
    \hline
    Selection & Average of Cosine Similarities\\
    \hline
       random  &  0.448\\
       similarity &  0.724\\
       tag-based  &  0.569 \\
       \hline
    \end{tabular}
    }
    \label{tab:average_similarity}
\end{table}

\subsection{Discussion}
This section discusses the precision-related findings.
\subsubsection{Task Misinterpretation with LLaMA 3.3}
With similarity setting, LLaMA 3.3 produced unexpected outputs in 207 cases (approximately 10\%).
These unexpected outputs exhibited two distinct patterns. First, some outputs regenerated background/causal factors and preventive measures for all five example cases already provided in the prompt, treating them as new inputs rather than reference examples. 
Second, other outputs summarized the five example cases into a single aggregated incident and generated only preventive measures for this artificial combined case, completely ignoring the actual input case requiring analysis (Table~\ref{tab:llama_ex}).

\subsubsection{Violence Filter Activation with GPT-4o}
With GPT-4o, 19 cases were blocked by the violence filter despite being standard medical reports. We hypothesize that extreme similarity amplified certain semantic features (e.g., patient harm terms), sharpening meanings and triggering the violence filter inappropriately.

\subsubsection{Stability and Advantages of Tag-Based Example Selection}
In contrast, in both the \textbf{random} and \textbf{tag-based} settings, GPT-4o and LLaMA 3.3 generated outputs for all cases. Unlike the similarity setting, GPT-4o showed stable generation without content-filter blocking, while maintaining high BERTScore and ROUGE-L precision in the \textbf{tag-based} setting. This supports its suitability for high-stakes domains such as healthcare, where high precision is critical.
Moreover, prior work \cite{skillbasedfewshotselectionincontext} suggests that selecting examples containing task-relevant skills—rather than purely high cosine similarity—can improve performance. Consistent with this, our \textbf{tag-based}, which uses shared dataset tags, increased BERTScore and ROUGE-L precision. 
This implies that tags function as human-interpretable proxies for skill-based subset selection, making them both effective and practical criteria for real-world deployment.

\section{future work}

Although the tag-based setting outperformed the random and similarity settings in terms of precision, two key directions warrant further investigation.

First, dynamic example selection within tag-based constraints should be explored. We used fixed example sets for reproducibility, but dynamically selecting examples for each input might yield even higher BERTScore and ROUGE-L precision, given content variation within categories.

Second, clinical validation with healthcare professionals is essential.
As shown in Table~\ref{output_example}, the model sometimes generates additional terms that are absent from the Reference Cause, which lowers the BERTScore and ROUGE-L precision score under automatic evaluation.
However, such additions are not necessarily errors, since medical incident analysis often admits multiple valid interpretations.
Therefore, evaluation by healthcare professionals would offer clinically grounded insights and is crucial before real-world deployment.


\begin{table*}[]
    \centering
    \caption{An example of unintended LlaMA output in the \textbf{similarity} setting, where the model summarizes all five input cases and generates only preventive measures.}
    \renewcommand{\arraystretch}{1.4}
    \begin{tabular}{|p{17cm}|}
         \hline
この文章は、医療における薬剤の処方や調剤に関するミスやトラブルの具体例と、それに対する改善策について述べたものです。以下に各ケースの要約と改善策をまとめます。\\
        (This text describes concrete examples of errors and troubles related to medication prescription and dispensing in healthcare, along with measures for improvement. Below is a summary of each case and its corresponding countermeasures.)\\
1. **薬剤の混同** \\
- 具体的内容: 薬剤の名前が似ていることによる混同。 \\
- 改善策: 処方箋や薬包装に注意深く確認する。薬剤名の類似性についてスタッフを教育する。\\
(1. **Medication Confusion** 
- Specific Issue: Confusion caused by similarities in drug names.\\
- Improvement Measure: Carefully check prescriptions and drug packaging. Educate staff about similarities in medication names.)\\
2. **退院時処方ミス** \\
- 具体的内容: 退院時に患者に与える薬剤の指示が不明確だったことによるミス。 \\
- 改善策: 退院時の処方に関する医師の指示を明確にする。カルテや口頭指示受けメモの確認を徹底する。\\
2. **Discharge Prescription Error**\\
- Specific Issue: Errors caused by unclear instructions regarding medications to be given at the time of patient discharge.\\
- Improvement Measure: Clarify physicians’ instructions for discharge prescriptions. Ensure thorough verification of medical records and handwritten notes from verbal instructions.\\
3. **内服薬の準備ミス** - 具体的内容: 内服薬の準備時に名前の記載をせずに投与したことによるミス。 - 改善策: 薬剤投与は全てを行って初めて確実な与薬であることを確認する。ひとつひとつの確認作業を徹底する。\\
3. **Oral Medication Preparation Error** \\
- Specific Issue: Error caused by administering oral medication without labeling the preparation.\\
- Improvement Measure: Ensure that all steps of medication administration are completed before confirming it as safely administered. Reinforce thorough verification at each step.\\
4. **調剤ミス**\\
- 具体的内容: 希望剤形の確認を怠ったことによるミス。 \\
- 改善策: 調剤者・監査者は処方箋上の薬剤名、用量、用法、処方日数の確認に加え、希望剤形の確認を徹底する。散剤秤量者は調剤すべき薬剤を見落とさない。\\
4. **Dispensing Error**\\
- Specific Issue: Error caused by failing to confirm the preferred dosage form.\\
- Improvement Measure: Both the dispenser and auditor must not only verify the drug name, dosage, administration method, and prescription duration on the prescription, but also confirm the preferred dosage form. Staff weighing powdered drugs must ensure that no prescribed medication is overlooked.\\
5. **オメプラール坐薬と五苓散座薬の交換ミス** \\
- 具体的内容: 薬袋の表示が間違っていたことによるミス。\\
- 改善策: 薬袋のラベルを正しく確認する。調剤および監査プロセスを強化し、薬剤の包装やラベリングに注意を払う。\\
\textit{(5. **Mix-up Between Omeprazole Suppository and Goreisan Suppository**}\\
\textit{- Specific Issue: Error caused by incorrect labeling on the medication bag.}\\
\textit{- Improvement Measure: Carefully verify medication bag labels. Strengthen the dispensing and auditing process, and pay close attention to drug packaging and labeling.)}\\
これらの改善策は、医療現場における薬剤関連ミスの防止と患者安全の向上に寄与します。\\
(By implementing these measures, medication-related errors in healthcare settings can be prevented, thereby improving patient safety.)\\
         \hline
    \end{tabular}
    \label{tab:llama_ex}
\end{table*}

\begin{table*}
    \centering
    \caption{Example generations by GPT-4o under zero-shot, similarity, random, and tag-based settings. English translation is shown in parentheses : The model outputs include information not present in the Reference Cause, which likely contributes to lower precision.
    In the similarity setting, elements not contained in the Reference Cause were introduced, such as problems of information transfer between outpatient and inpatient care (e.g., was not properly communicated). Likewise, in the random setting, unnecessary content was generated, including references to medical records such as the “カルテ,” which were not mentioned in the reference. These additions likely contributed to the decrease in precision.}
    {
    \renewcommand{\arraystretch}{1.4}

    \begin{tabular}{|l|}
    \hline
    input\\
    \hline
    入院前に外来でアルダクトンＡ１錠（朝）の処方があり、患者は内服をしていた。入院後に主治医がそれに気付かず、同薬を同量処方をした。\\
    薬局で調剤時には、薬歴をチェックするようになっているが、日数の確認不十分で処方され、重複投与を７日間してしまった。\\
    (Before admission, the outpatient clinic had prescribed Aldactone A, one tablet each morning, and the patient had been taking it. After admission, \\
    the attending physician failed to recognize this and prescribed the same medication at the same dose. Although the pharmacy is supposed to  check\\
    the medication history at the time of dispensing, the duration was not adequately verified, the prescription was filled, and the patient received\\
    duplicate therapy for seven days.)\\
\hline
Reference Cause\\
\hline
入院後、入院前の服用薬をチェックリストで確認しているが、当院処方のものは、薬歴にあるため薬名と個数のみの確認であった。調剤後に\\再度他の薬剤師が監査を実施しているが、処方箋との確認であり、薬歴を再度確認は出来ていない。今回は、新人薬剤師で多忙な時間帯で\\あったため薬歴のチェックが不十分で、重複していることに気付かなかったためにヒヤリ・ハットが発生した。\\
(After admission, we verify pre-admission medications using a checklist. However, for medications prescribed by our hospital, since they are already\\
listed in the medication history, only the drug name and quantity were confirmed. After dispensing, another pharmacist conducts a secondary audit,\\
but this is a check against the prescription only, and the medication history is not reviewed again. In this case, because it was a busy period and\\
the pharmacist was a junior staff member, the review of the medication history was insufficient, the duplication went unnoticed, and a near-miss \\incident occurred.)\\
\hline
\hline
\textit{output / zero-shot} : Prec : 0.694\\
\hline
1. **情報共有の不足**：外来と入院の間で患者の薬歴情報が十分に共有されていなかったため、主治医が既に処方されている薬を把握でき\\なかった。\\
2. **薬歴チェックの不徹底**：薬局での調剤時に薬歴の確認が不十分であり、重複投与のリスクが見逃された。\\
3. **システムの不備**：電子カルテや薬歴管理システムが十分に連携しておらず、患者の全体的な薬物治療状況が一元管理されていなかった。\\
(1. **Lack of Information Sharing**: Medication history information was not sufficiently shared between the outpatient and inpatient settings,so\\
the attending physician was not aware of drugs that had already been prescribed.\\
2. **Inadequate Review of Medication History**: During dispensing in the pharmacy, the medication history was not thoroughly checked, and the\\
risk of duplicate therapy was overlooked.\\
3. **System Deficiency**: The electronic medical record and medication management systems were not adequately integrated, so the patient's \\
overall medication regimen was not managed in a unified manner.)\\
\hline
\textit{output / similarity} : Prec. : 0.784\\
\hline
入院前の外来での処方内容が入院後の主治医に適切に伝達されていなかった。また、薬局での薬歴チェックが不十分であり、日数の確認が行わ\\れていなかったため、重複投与が発生した。\\
(The medication prescribed at the outpatient clinic before admission was not properly communicated to the attending physician after admission. In \\
addition, the medication history check at the pharmacy was insufficient, and the number of days was not verified, resulting in duplicate administration.)\\
\hline
\textit{output / random} : Prec : 0.699\\
\hline
入院前の外来での処方内容が入院後のカルテに正確に反映されていなかった。薬局での調剤時に薬歴のチェックが不十分であり、重複投与の\\確認が行われなかった。\\
(The outpatient prescription prior to admission was not accurately reflected in the inpatient medical record. During dispensing at the pharmacy, \\
the medication history check was insufficient, and duplicate administration was not identified.)\\
\hline
\textbf{\textit{output / tag-based}} : Prec. : \textbf{0.809}\\
\hline
入院前の外来処方が確認されず、入院後の処方と重複してしまった。薬歴のチェックが不十分であり、日数の確認が徹底されていなかった。\\
(The outpatient prescription prior to admission was not verified, and it overlapped with the inpatient prescription. The medication history check was \\
insufficient, and confirmation of the prescribed days was not thoroughly conducted.)\\
\hline

\end{tabular}
}
\label{output_example}
\end{table*}

\section{conclusion}
In order to develop the method of automatically generating the background/causal factors and preventive measures of each medical incident, in this paper, we proposed a novel example
selection method called ”tag-based example selection".
Experimental evaluation using a real textual dataset demonstrates that our approach outperforms the other baselines in BERTScore and ROUGE-L precision score in a few-shot learning setting with GPT-4o and LLaMA 3.3.
While BERTScore and ROUGE-L precision were the primary focus, experimental results also revealed that this approach contributed to more stable generation behavior, highlighting an additional benefit of the method.
Future work should further explore hybrid selection strategies that balance semantic relevance with task-specific diversity, aiming to enhance the robustness and trustworthiness of LLMs in medical environments.

\section*{Acknowledgment}
This work was supported in part by JST-PRESTO Grant Number JPMJPR2267 and JST-CREST Grant Number PMJCR2564, Japan.
%






\bibliographystyle{IEEEtran}
\bibliography{ref}

@inproceedings{skillbasedfewshotselectionincontext,
    title = "Skill-Based Few-Shot Selection for In-Context Learning",
    author = "An, Shengnan  and
      Zhou, Bo  and
      Lin, Zeqi  and
      Fu, Qiang  and
      Chen, Bei  and
      Zheng, Nanning  and
      Chen, Weizhu  and
      Lou, Jian-Guang",
    editor = "Bouamor, Houda  and
      Pino, Juan  and
      Bali, Kalika",
    booktitle = "Proceedings of the 2023 Conference on Empirical Methods in Natural Language Processing",
    month = dec,
    year = "2023",
    address = "Singapore",
    publisher = "Association for Computational Linguistics",
    url = "https://aclanthology.org/2023.emnlp-main.831/",
    doi = "10.18653/v1/2023.emnlp-main.831",
    pages = "13472--13492",
}

@misc{random,
      title={Use Random Selection for Now: Investigation of Few-Shot Selection Strategies in LLM-based Text Augmentation for Classification}, 
      author={Jan Cegin and Branislav Pecher and Jakub Simko and Ivan Srba and Maria Bielikova and Peter Brusilovsky},
      year={2024},
      eprint={2410.10756},
      archivePrefix={arXiv},
      url={https://arxiv.org/abs/2410.10756}, 
}

@article{chen2024unleashingpotentialpromptengineering,
  title = {Unleashing the Potential of Prompt Engineering for Large Language Models},
  author = {Chen, Banghao and Zhang, Zhaofeng and Langrené, Nicolas and Zhu, Shengxin},
  date = {2025},
  journaltitle = {Patterns},
  volume = {6},
  number = {6},
  pages = {101260},
  issn = {2666-3899},
  doi = {10.1016/j.patter.2025.101260},
  url = {https://www.sciencedirect.com/science/article/pii/S2666389925001084},
}

@misc{meditron,
      title={MEDITRON-70B: Scaling Medical Pretraining for Large Language Models}, 
      author={Zeming Chen and Alejandro Hernández Cano and Angelika Romanou and Antoine Bonnet and Kyle Matoba and Francesco Salvi and Matteo Pagliardini and Simin Fan and Andreas Köpf and Amirkeivan Mohtashami and Alexandre Sallinen and Alireza Sakhaeirad and Vinitra Swamy and Igor Krawczuk and Deniz Bayazit and Axel Marmet and Syrielle Montariol and Mary-Anne Hartley and Martin Jaggi and Antoine Bosselut},
      year={2023},
      eprint={2311.16079},
      archivePrefix={arXiv},
      url={https://arxiv.org/abs/2311.16079}, 
}

@article{communicate,
  title = {The Future Landscape of Large Language Models in Medicine},
  author = {Clusmann, Jan and Kolbinger, Fiona R. and Muti, Hannah Sophie and Carrero, Zunamys I. and Eckardt, Jan-Niklas and Laleh, Narmin Ghaffari and Löffler, Chiara Maria Lavinia and Schwarzkopf, Sophie-Caroline and Unger, Michaela and Veldhuizen, Gregory P. and Wagner, Sophia J. and Kather, Jakob Nikolas},
  date = {2023-10-10},
  journaltitle = {Communications Medicine},
  shortjournal = {Communications Medicine},
  volume = {3},
  number = {1},
  pages = {141},
  issn = {2730-664X},
  doi = {10.1038/s43856-023-00370-1},
  url = {https://doi.org/10.1038/s43856-023-00370-1},
}

@article{summarization,
  title = {Adapted Large Language Models Can Outperform Medical Experts in Clinical Text Summarization},
  author = {Van Veen, Dave and Van Uden, Cara and Blankemeier, Louis and Delbrouck, Jean-Benoit and Aali, Asad and Bluethgen, Christian and Pareek, Anuj and Polacin, Malgorzata and Reis, Eduardo Pontes and Seehofnerová, Anna and Rohatgi, Nidhi and Hosamani, Poonam and Collins, William and Ahuja, Neera and Langlotz, Curtis P. and Hom, Jason and Gatidis, Sergios and Pauly, John and Chaudhari, Akshay S.},
  date = {2024-04-01},
  journaltitle = {Nature Medicine},
  shortjournal = {Nature Medicine},
  volume = {30},
  number = {4},
  pages = {1134--1142},
  issn = {1546-170X},
  doi = {10.1038/s41591-024-02855-5},
  url = {https://doi.org/10.1038/s41591-024-02855-5},
}

@misc{bert_model,
      title={BERT: Pre-training of Deep Bidirectional Transformers for Language Understanding}, 
      author={Jacob Devlin and Ming-Wei Chang and Kenton Lee and Kristina Toutanova},
      year={2019},
      eprint={1810.04805},
      archivePrefix={arXiv},
      url={https://arxiv.org/abs/1810.04805}, 
}

@misc{medqa,
      title={What Disease does this Patient Have? A Large-scale Open Domain Question Answering Dataset from Medical Exams}, 
      author={Di Jin and Eileen Pan and Nassim Oufattole and Wei-Hung Weng and Hanyi Fang and Peter Szolovits},
      year={2020},
      eprint={2009.13081},
      archivePrefix={arXiv},
      url={https://arxiv.org/abs/2009.13081}, 
}

@misc{pubmedqa,
      title={PubMedQA: A Dataset for Biomedical Research Question Answering}, 
      author={Qiao Jin and Bhuwan Dhingra and Zhengping Liu and William W. Cohen and Xinghua Lu},
      year={2019},
      eprint={1909.06146},
      archivePrefix={arXiv},
      url={https://arxiv.org/abs/1909.06146}, 
}

@misc{igakuqa,
      title={Evaluating GPT-4 and ChatGPT on Japanese Medical Licensing Examinations}, 
      author={Jungo Kasai and Yuhei Kasai and Keisuke Sakaguchi and Yutaro Yamada and Dragomir Radev},
      year={2023},
      eprint={2303.18027},
      archivePrefix={arXiv},
      url={https://arxiv.org/abs/2303.18027}, 
}

@misc{hallucination,
      title={Medical Hallucinations in Foundation Models and Their Impact on Healthcare}, 
      author={Yubin Kim and Hyewon Jeong and Shan Chen and Shuyue Stella Li and Mingyu Lu and Kumail Alhamoud and Jimin Mun and Cristina Grau and Minseok Jung and Rodrigo Gameiro and Lizhou Fan and Eugene Park and Tristan Lin and Joonsik Yoon and Wonjin Yoon and Maarten Sap and Yulia Tsvetkov and Paul Liang and Xuhai Xu and Xin Liu and Daniel McDuff and Hyeonhoon Lee and Hae Won Park and Samir Tulebaev and Cynthia Breazeal},
      year={2025},
      eprint={2503.05777},
      url={https://arxiv.org/abs/2503.05777}, 
}

@article{patientreports,
author = {Kripalani, Sunil and Lefevre, Frank and Phillips, Christopher and Williams, Mark and Basaviah, Preetha and Baker, David},
year = {2007},
month = {03},
pages = {831-41},
title = {Deficits in Communication and Information Transfer Between Hospital-Based and Primary Care Physicians},
volume = {297},
journal = {JAMA : the journal of the American Medical Association},
doi = {10.1001/jama.297.8.831}
}

@inproceedings{gpt3,
    title = "What Makes Good In-Context Examples for {GPT}-3?",
    author = "Liu, Jiachang  and
      Shen, Dinghan  and
      Zhang, Yizhe  and
      Dolan, Bill  and
      Carin, Lawrence  and
      Chen, Weizhu",
    editor = "Agirre, Eneko  and
      Apidianaki, Marianna  and
      Vuli{\'c}, Ivan",
    booktitle = "Proceedings of Deep Learning Inside Out (DeeLIO 2022): The 3rd Workshop on Knowledge Extraction and Integration for Deep Learning Architectures",
    month = may,
    year = "2022",
    address = "Dublin, Ireland and Online",
    publisher = "Association for Computational Linguistics",
    url = "https://aclanthology.org/2022.deelio-1.10/",
    doi = "10.18653/v1/2022.deelio-1.10",
    pages = "100--114",
}

@misc{automaticcombinationsampleselection,
      title={Automatic Combination of Sample Selection Strategies for Few-Shot Learning}, 
      author={Branislav Pecher and Ivan Srba and Maria Bielikova and Joaquin Vanschoren},
      year={2024},
      eprint={2402.03038},
      archivePrefix={arXiv},
      url={https://arxiv.org/abs/2402.03038}, 
}

@article{towards,
  title = {Towards Building Multilingual Language Model for Medicine},
  author = {Qiu, Pengcheng and Wu, Chaoyi and Zhang, Xiaoman and Lin, Weixiong and Wang, Haicheng and Zhang, Ya and Wang, Yanfeng and Xie, Weidi},
  date = {2024-09-27},
  journaltitle = {Nature Communications},
  shortjournal = {Nature Communications},
  volume = {15},
  number = {1},
  pages = {8384},
  issn = {2041-1723},
  doi = {10.1038/s41467-024-52417-z},
  url = {https://doi.org/10.1038/s41467-024-52417-z},
}

@misc{medgemini,
      title={Capabilities of Gemini Models in Medicine}, 
      author={Khaled Saab and Tao Tu and Wei-Hung Weng and Ryutaro Tanno and David Stutz and Ellery Wulczyn and Fan Zhang and Tim Strother and Chunjong Park and Elahe Vedadi and Juanma Zambrano Chaves and Szu-Yeu Hu and Mike Schaekermann and Aishwarya Kamath and Yong Cheng and David G. T. Barrett and Cathy Cheung and Basil Mustafa and Anil Palepu and Daniel McDuff and Le Hou and Tomer Golany and Luyang Liu and Jean-baptiste Alayrac and Neil Houlsby and Nenad Tomasev and Jan Freyberg and Charles Lau and Jonas Kemp and Jeremy Lai and Shekoofeh Azizi and Kimberly Kanada and SiWai Man and Kavita Kulkarni and Ruoxi Sun and Siamak Shakeri and Luheng He and Ben Caine and Albert Webson and Natasha Latysheva and Melvin Johnson and Philip Mansfield and Jian Lu and Ehud Rivlin and Jesper Anderson and Bradley Green and Renee Wong and Jonathan Krause and Jonathon Shlens and Ewa Dominowska and S. M. Ali Eslami and Katherine Chou and Claire Cui and Oriol Vinyals and Koray Kavukcuoglu and James Manyika and Jeff Dean and Demis Hassabis and Yossi Matias and Dale Webster and Joelle Barral and Greg Corrado and Christopher Semturs and S. Sara Mahdavi and Juraj Gottweis and Alan Karthikesalingam and Vivek Natarajan},
      year={2024},
      eprint={2404.18416},
      archivePrefix={arXiv},
      url={https://arxiv.org/abs/2404.18416}, 
}

@article{multimedqa,
  title = {Large Language Models Encode Clinical Knowledge},
  author = {Singhal, Karan and Azizi, Shekoofeh and Tu, Tao and Mahdavi, S. Sara and Wei, Jason and Chung, Hyung Won and Scales, Nathan and Tanwani, Ajay and Cole-Lewis, Heather and Pfohl, Stephen and Payne, Perry and Seneviratne, Martin and Gamble, Paul and Kelly, Chris and Babiker, Abubakr and Schärli, Nathanael and Chowdhery, Aakanksha and Mansfield, Philip and Demner-Fushman, Dina and Agüera y Arcas, Blaise and Webster, Dale and Corrado, Greg S. and Matias, Yossi and Chou, Katherine and Gottweis, Juraj and Tomasev, Nenad and Liu, Yun and Rajkomar, Alvin and Barral, Joelle and Semturs, Christopher and Karthikesalingam, Alan and Natarajan, Vivek},
  date = {2023-08-01},
  journaltitle = {Nature},
  shortjournal = {Nature},
  volume = {620},
  number = {7972},
  pages = {172--180},
  issn = {1476-4687},
  doi = {10.1038/s41586-023-06291-2},
  url = {https://doi.org/10.1038/s41586-023-06291-2},
}

@article{medpalm2,
  title = {Toward Expert-Level Medical Question Answering with Large Language Models},
  author = {Singhal, Karan and Tu, Tao and Gottweis, Juraj and Sayres, Rory and Wulczyn, Ellery and Amin, Mohamed and Hou, Le and Clark, Kevin and Pfohl, Stephen R. and Cole-Lewis, Heather and Neal, Darlene and Rashid, Qazi Mamunur and Schaekermann, Mike and Wang, Amy and Dash, Dev and Chen, Jonathan H. and Shah, Nigam H. and Lachgar, Sami and Mansfield, Philip Andrew and Prakash, Sushant and Green, Bradley and Dominowska, Ewa and Agüera y Arcas, Blaise and Tomašev, Nenad and Liu, Yun and Wong, Renee and Semturs, Christopher and Mahdavi, S. Sara and Barral, Joelle K. and Webster, Dale R. and Corrado, Greg S. and Matias, Yossi and Azizi, Shekoofeh and Karthikesalingam, Alan and Natarajan, Vivek},
  date = {2025-03-01},
  journaltitle = {Nature Medicine},
  shortjournal = {Nature Medicine},
  volume = {31},
  number = {3},
  pages = {943--950},
  issn = {1546-170X},
  doi = {10.1038/s41591-024-03423-7},
  url = {https://doi.org/10.1038/s41591-024-03423-7},
}

@misc{bertscore,
      title={BERTScore: Evaluating Text Generation with BERT}, 
      author={Tianyi Zhang and Varsha Kishore and Felix Wu and Kilian Q. Weinberger and Yoav Artzi},
      year={2020},
      eprint={1904.09675},
      archivePrefix={arXiv},
      url={https://arxiv.org/abs/1904.09675}, 
}

@inproceedings{few-shot,
  title = {Language Models Are Few-Shot Learners},
  booktitle = {Advances in Neural Information Processing Systems},
  author = {Brown, Tom and Mann, Benjamin and Ryder, Nick and Subbiah, Melanie and Kaplan, Jared D and Dhariwal, Prafulla and Neelakantan, Arvind and Shyam, Pranav and Sastry, Girish and Askell, Amanda and Agarwal, Sandhini and Herbert-Voss, Ariel and Krueger, Gretchen and Henighan, Tom and Child, Rewon and Ramesh, Aditya and Ziegler, Daniel and Wu, Jeffrey and Winter, Clemens and Hesse, Chris and Chen, Mark and Sigler, Eric and Litwin, Mateusz and Gray, Scott and Chess, Benjamin and Clark, Jack and Berner, Christopher and McCandlish, Sam and Radford, Alec and Sutskever, Ilya and Amodei, Dario},
  editor = {Larochelle, H. and Ranzato, M. and Hadsell, R. and Balcan, M.F. and Lin, H.},
  date = {2020},
  volume = {33},
  pages = {1877--1901},
  publisher = {Curran Associates, Inc.},
  url = {https://proceedings.neurips.cc/paper_files/paper/2020/file/1457c0d6bfcb4967418bfb8ac142f64a-Paper.pdf}
}

@inproceedings{rouge,
    title = "{ROUGE}: A Package for Automatic Evaluation of Summaries",
    author = "Lin, Chin-Yew",
    booktitle = "Text Summarization Branches Out",
    month = jul,
    year = "2004",
    address = "Barcelona, Spain",
    publisher = "Association for Computational Linguistics",
    url = "https://aclanthology.org/W04-1013/",
    pages = "74--81"
}

\vspace{12pt}

\clearpage
\appendices

\section{Mapping of Fine-Grained Labels to Broad Categories}
In this study, the data were classified based on the “descriptive information” tags contained in the dataset. For categories with a small amount of data, fine-grained labels were consolidated, resulting in a total of 18 categories. Part of this mapping is shown in Table \ref{map}.


\begin{table}[h]
    \centering
    \caption{Mapping of Fine-Grained Labels to Broad Categories}
        \renewcommand{\arraystretch}{1.2}
        \label{map}
    \begin{tabular}{p{4cm}|p{4cm}}
\hline
\textbf{Fine-Grained Labels} & \textbf{Broad Category} \\
\hline
Medications;\newline
Other Medications 
& Medications \\
\hline
Drain Insertion and Management;\newline
Nasogastric Tube;\newline
Gastrostomy and Enterostomy Tube;\newline
Enema 
& Drain Insertion and Management \\
\hline
Clinical Tests;\newline
Laboratory Tests 
& (Clinical) Laboratory Tests \\
\hline
Laterality Error;\newline
Other Site Errors;\newline
Non-procedural Other Site Errors 
& Left-Right (Body Part) Confusion \\
\hline
    \end{tabular}
\end{table}




\section{Prompt Design}
The prompts used in this study are shown in \ref{tab:prompt}.
In addition, we applied essentially the same prompt design in the random, similarity, and tag-based approaches. In contrast, under the zero-shot condition, we adopted a setting where no examples were provided.

\begin{table}[h]
    \centering
    \caption{
The placeholders {details of medical accident 1}, {background/causal factors 1}, and {preventive measures 1} are filled with the selected examples.}
    \resizebox{0.49\textwidth}{!}{
    \renewcommand{\arraystretch}{1.2}

    \begin{tabular}{|l|}
    \hline
The following presents the details of a medical incident,\\
along with its background/causal factors and preventive\\
measures. Please refer to the examples below and \\generate only the background/causal factors and preventive
\\measures.\\
Specifics: \{details of medical accident 1\}\\
Background/causal factors: \{background/causal factors 1\} \\
Preventive measures: \{preventive measures 1\}\\
$\vdots$ \\
(repeat for 4 examples)\\
$\vdots$
\\
Specifics: \{INPUT\}\\
Background/causal factors: \\
Preventive measures:\\
\hline
    \end{tabular}
    }
    \label{tab:prompt}

\end{table}

\end{document}